\documentclass{article}

\usepackage{arxiv}

\usepackage[utf8]{inputenc} % allow utf-8 input
\usepackage[T1]{fontenc}    % use 8-bit T1 fonts
\usepackage{hyperref}       % hyperlinks
\usepackage{url}            % simple URL typesetting
\usepackage{booktabs}       % professional-quality tables
\usepackage{amsmath,amssymb,amsfonts}       % blackboard math symbols
\usepackage{nicefrac}       % compact symbols for 1/2, etc.
\usepackage{microtype}      % microtypography
\usepackage{lipsum}
\usepackage{graphicx}
\graphicspath{ {./images/} }

\title{GLaRE: A Graph-based Landmark Region Embedding Network for Emotion Recognition}

\author{
 Debasis Maji \\
  Department of Computer \& System Sciences\\
  Visva-Bharati\\
  India, 731235 \\
  \texttt{youdebasis@gmail.com} \\
   \And
 Debaditya Barman \\
  Department of Computer \& System Sciences\\
  Visva-Bharati\\
  India, 731235 \\
  \texttt{debadityabarman@gmail.com} \\
}

\begin{document}
\maketitle
\begin{abstract}
Facial expression recognition (FER) is a crucial task in computer vision with wide range of applications including human–computer interaction, surveillance, and assistive technologies. However, challenges such as occlusion, expression variability, and lack of interpretability hinder the performance of traditional FER systems. Graph Neural Networks (GNNs) offer a powerful alternative by modeling relational dependencies between facial landmarks, enabling structured and interpretable learning. In this paper, we propose GLaRE, a novel Graph-based Landmark Region Embedding network for emotion recognition. Facial landmarks are extracted using 3D facial alignment, and a quotient graph is constructed via hierarchical coarsening to preserve spatial structure while reducing complexity. Our method achieves $\sim$64.89\% accuracy on AffectNet\footnote{https://www.mohammadmahoor.com/pages/databases/affectnet/} and $\sim$94.24\% on FERG\footnote{https://grail.cs.washington.edu/projects/deepexpr/ferg-2d-db.html}, outperforming several existing baselines. Additionally, ablation studies have demonstrated that region-level embeddings from quotient graphs have contributed to improved prediction performance.
\end{abstract}

% keywords can be removed
\keywords{Facial Expression Recognition \and Graph Neural Network \and Deep learning \and Human Computer Interaction.}

\section{Introduction}
Emotion recognition through facial expressions is one of the essential components of artificial intelligence (AI), particularly in the area of computer vision and human-computer interaction (HCI). Among the various modalities for emotion recognition, such as audio, video, and bio-signals, facial gestures offer a direct and intuitive cue to human emotions. Facial landmarks, which abstract away identifiable facial details, offer an efficient and privacy-preserving alternative to the processing of the entire image of the face by capturing only key facial positions. Their subtle movements are strongly associated with different affective states, making them well-suited for real-time applications in smart devices~\cite{pampouchidou2020automated, srinivasan2024iot}, driver monitoring systems~\cite{gao2014detecting, kaushik2024deep}, e-learning~\cite{fang2020computer}, and immersive gaming environments~\cite{marin2020emotion}. This has resulted in development of Facial Landmark based Emotion Recognition, which holds promise for both accuracy and resource efficiency.
\begin{figure}[ht!]
    \centering
    \includegraphics[width=\textwidth]{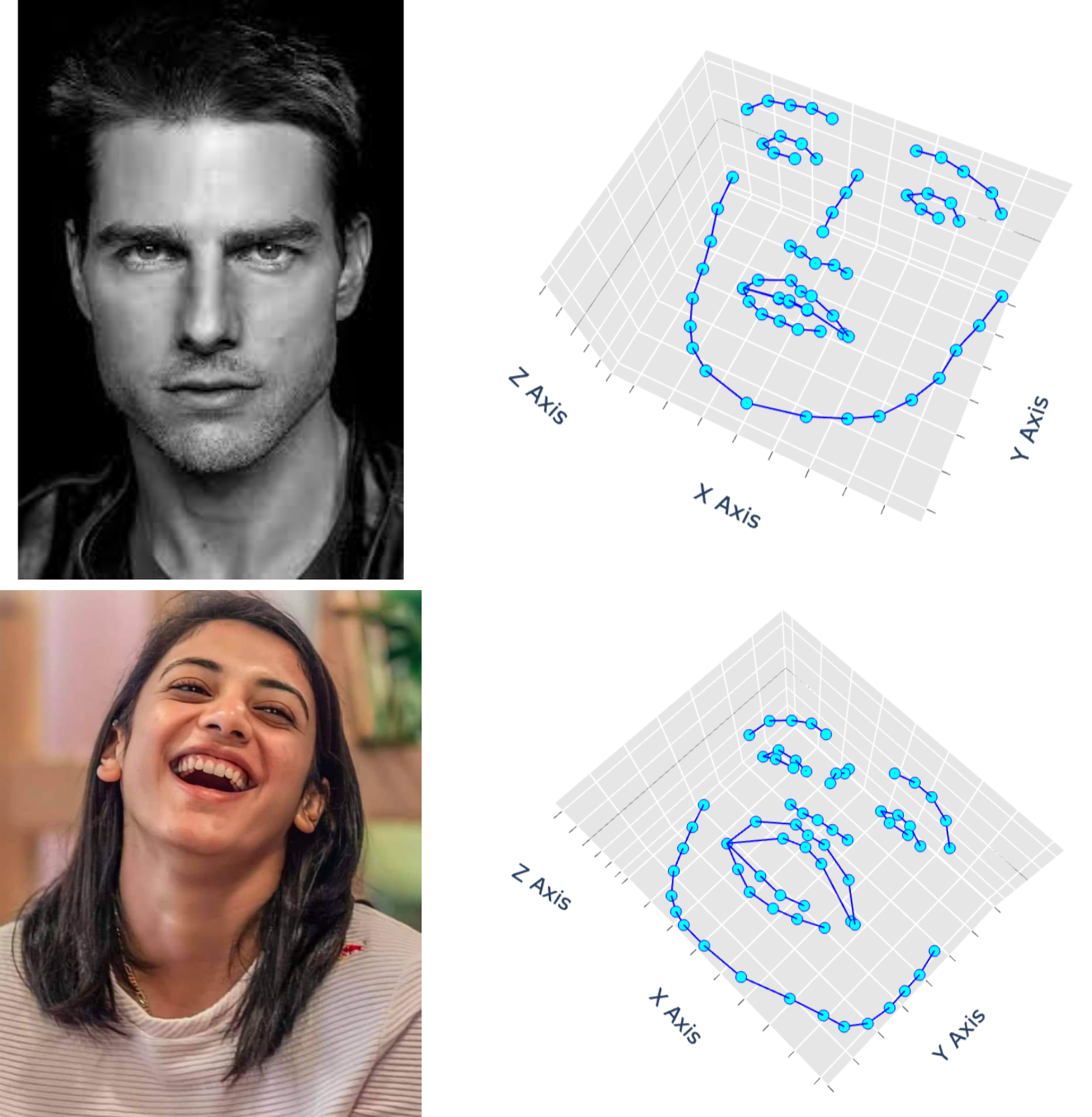}
    \caption{Visual comparison of facial expression images and their corresponding 3D facial landmarks.}
    \label{fig:intro}
\end{figure}

Although facial landmarks~\cite{chen2021residual} provide semantically rich and privacy preserving indicators for Facial Expression Recognition (FER), they have remained underutilized due to the lack of effective methods to model their spatial and temporal dependencies. While early studies have demonstrated the effectiveness of combining landmarks with appearance features or integrating them in multimodal pipelines (i.e., with EEG or video data)~\cite{kuo2018compact,hossain2019emotion,zhang2018spatial} , standalone landmark based methods require advanced geometric reasoning to reach full potential. The advancement of geometric deep learning and graph neural networks (GNNs) provides further opportunities to effectively utilize facial landmarks for FER.

Convolutional neural networks (CNNs)~\cite{liu2014deeply,meng2017identity, yu2025cross} have achieved notable success in FER by learning local patterns from facial images. However, they often fall short in capturing the geometric relationships among facial landmarks. Most CNN-based methods focus on extracting deep appearance features, overlooking the underlying spatial structure of the face, which limits generalization and interpretability.

To overcome these limitations, GNNs~\cite{huang2025modeling, liu2024descriptive, ai2024gcn} have emerged as a powerful alternative. Unlike CNNs, GNNs naturally operate on non-Euclidean data such as facial landmarks, modeling both the local and global structural relationships among keypoints. In the context of FER, each facial landmark is modeled as a node in a facial landmark graph. Edges in the graph encode anatomical or spatial connections, enabling the model to capture physiologically meaningful interactions. This facial landmark graph formulation enhances representation and generalization while improving robustness and interpretability by aligning with human cognitive mechanisms. Fig.~\ref{fig:intro} illustrates two example of faces and their corresponding facial landmark graphs, highlighting how expression information can be structurally encoded. The key contributions of this work are as follows:
\begin{itemize}
    \item We have proposed a novel graph-based model, GLaRE (Graph-based Landmark Region Embedding Network), which effectively captures both local and global geometric relationships among facial landmarks for emotion recognition.
    \item We have introduced a hierarchical quotient graph coarsening mechanism that has significantly reduced computational complexity while preserving essential facial structural information, thereby rendering the proposed model lightweight and well-suited for real-time facial emotion recognition tasks.
\end{itemize}

\section{Problem Statement}
A facial image can be represented as a landmark graph $G= (V,E)$, where $V = \{v_{1},v_{2}, \cdots, v_{n}\}$ and $E \subseteq V \times V$ is the set of undirected edges connecting anatomically adjacent landmarks, and each node $v_{i} \in V$ is associated with a spatial coordinate feature vector $x_{i} \in \mathbb{R}^{d}$, derived using a landmark detector $\Phi$. Thus, the graph $G = (V, E, X)$ captures the facial structure, where $X \in \mathbb{R}^{d}$ is the matrix of node features. To incorporate anatomical hierarchy, the facial graph is partitioned into $R$ predefined facial regions.
\begin{align}
    V = V^{(1)} \cup V^{(2)} \cup \cdots \cup V^{(r)} \text{, where } V^{(r)} \subset V
\end{align}
Each facial graph $G_{i} = (V_{i}, E_{i}, X_{i})$ in the dataset is associated with a ground-truth emotion label $y_{i} \in \mathcal{Y}$ where $\mathcal{Y} = \{1,2,\cdots,C\}$
denotes the set of $C$ discrete expression classes. The objective is to learn the following mapping that can accurately predict the corresponding expression label from the structural and spatial properties of the facial graph.
\begin{align}
    f:G_{i} \rightarrow y_{i}
\end{align}
 This defines a graph classification problem, where each input graph encodes a face and its configuration, and the output is a categorical expression.
\section{Proposed Method}
\subsection{Graph Construction from Facial Images}
To prepare the input image for graph-based learning, a three-stage preprocessing pipeline has been adopted. 

First, 3D facial landmarks have been extracted from images using the Facial Alignment Network (FAN)~\cite{bulat2017far}. Each image has been processed to identify a fixed set of landmark points, thereby capturing the geometric structure of key facial regions such as the eyes, nose, mouth, and jawline. These landmark coordinates \(\{\mathbf{p}_i \in \mathbb{R}^3\}_{i=1}^N\) have served as the geometric basis for node initialization.

Second, appearance features have been extracted using a pretrained ResNet-18 model~\cite{he2016deep}, where the global average pooling and fully connected layers have been excluded to retain spatial feature maps for bilinear interpolation at landmark positions. The sampled descriptors have then been reduced to $f$ dimensions using principal component analysis (PCA). These appearance features have been concatenated with the normalized 3D landmark coordinates, yielding node features.
\begin{align}
    \mathbf{x}_i = \big[\, \hat{\mathbf{p}}_i \;\|\; \mathbf{a}_i \,\big] \in \mathbb{R}^{f+3}, \quad i=1,\dots,N
\end{align}
where \(\hat{\mathbf{p}}_i \in \mathbb{R}^3\) denotes normalized coordinates and \(\mathbf{a}_i \in \mathbb{R}^{f}\) denotes the reduced appearance descriptor.

Finally, a fine-level graph \(\mathcal{G}_f = (V_f, E_f)\) has been constructed, where each node \(v_i \in V_f\) corresponds to a landmark with feature vector \(\mathbf{x}_i\). Edges have been established using a \(k\)-nearest neighbor (kNN) criterion in Euclidean space:
\begin{align}
    E_f = \big\{ (v_i, v_j) \;|\; v_j \in \text{kNN}(v_i; \mathbf{p}_i) \big\}.
\end{align}
This procedure has ensured that each landmark is connected to its local neighborhood, thereby encoding fine-grained spatial relationships among facial components. Together, the node features and adjacency structure have provided a rich representation of facial topology. %, enabling context-aware feature extraction in subsequent learning stages.

\begin{figure}[ht!]
    \centering
    \includegraphics[width=\textwidth]{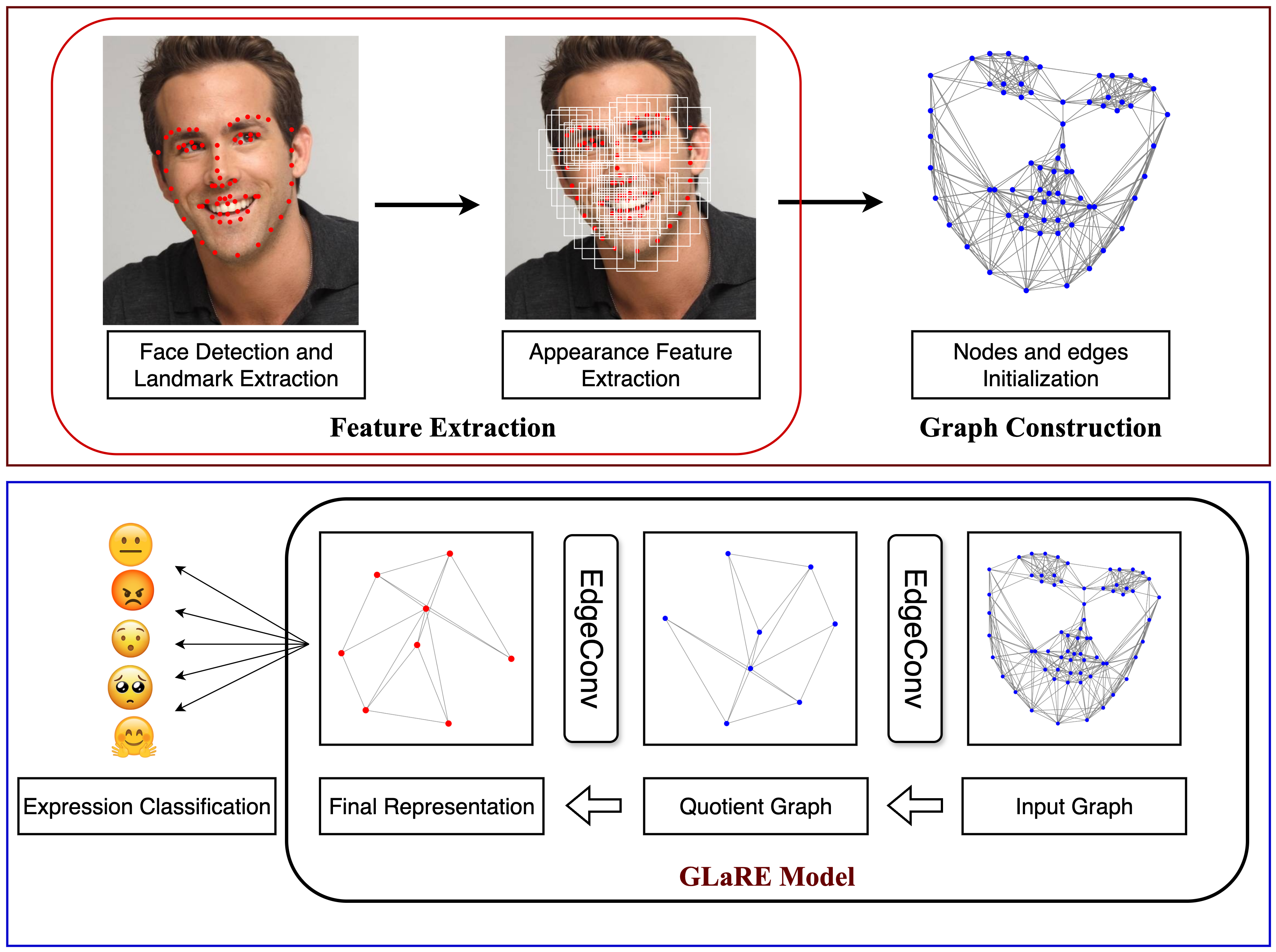}
    \caption{Schematic diagram of GLaRE model for facial expression recognition.}
    \label{fig:proposed_model}
\end{figure}

As illustrated in Fig.~\ref{fig:proposed_model}, the proposed GLaRE model has followed a two-level graph-based architecture that incorporates equivariant message passing. The architecture has been organized into three key stages: (a) fine-grained node embeddings have been computed from landmark graphs using equivariant GNN layers, (b) a quotient graph has been constructed to compress landmark-level information into region-level nodes, and (c) region embeddings have been further processed through equivariant GNN layers to produce the final global embedding. The subsequent subsections present each stage of the architecture in detail.
\subsection{Fine-Level Node Embedding using EdgeConv}
Given the fine-grained landmark graph, where each node corresponds to a facial landmark and edges encode spatial proximity, node embeddings are computed using an EdgeConv-based message passing scheme~\cite{wang2019dynamic}. In this scheme, the embedding of each node $i$ is updated by aggregating messages from its neighbors $j \in \mathcal{N}(i)$ using max aggregation, as defined in \eqref{eq:edgeconv}:

\begin{equation}
\mathbf{h}_i^{(l+1)} = \max_{j \in \mathcal{N}(i)} \phi\left(\mathbf{h}_i^{(l)}, \mathbf{h}_j^{(l)} - \mathbf{h}_i^{(l)}\right)
\label{eq:edgeconv}
\end{equation}

Here, $\mathbf{h}_i^{(l)}$ and $\mathbf{h}_j^{(l)}$ represent the feature vectors of the target and neighboring nodes at layer $l$, respectively. The function $\phi$, implemented as a two-layer multi-layer perceptron (MLP) with ReLU activation, transforms the concatenated source and relative features ($[\mathbf{h}_i^{(l)}, \mathbf{h}_j^{(l)} - \mathbf{h}_i^{(l)}]$) into informative edge-level messages. This formulation captures local differences in landmark configurations and appearance. Moreover, it enhances the representation of facial expression patterns for subsequent processing.

\subsection{Region-Level Embedding via Quotient Graph}
To capture higher-level facial structures critical for FER, a quotient graph is introduced to model regional patterns by aggregating landmark nodes into coarse-grained regions. This approach reduces computational complexity, enhances robustness to local landmark variations, and enables hierarchical processing of local and global facial features, improving the model's ability to distinguish expressions.

\begin{figure}[ht!]
    \centering
    \includegraphics[width=\textwidth]{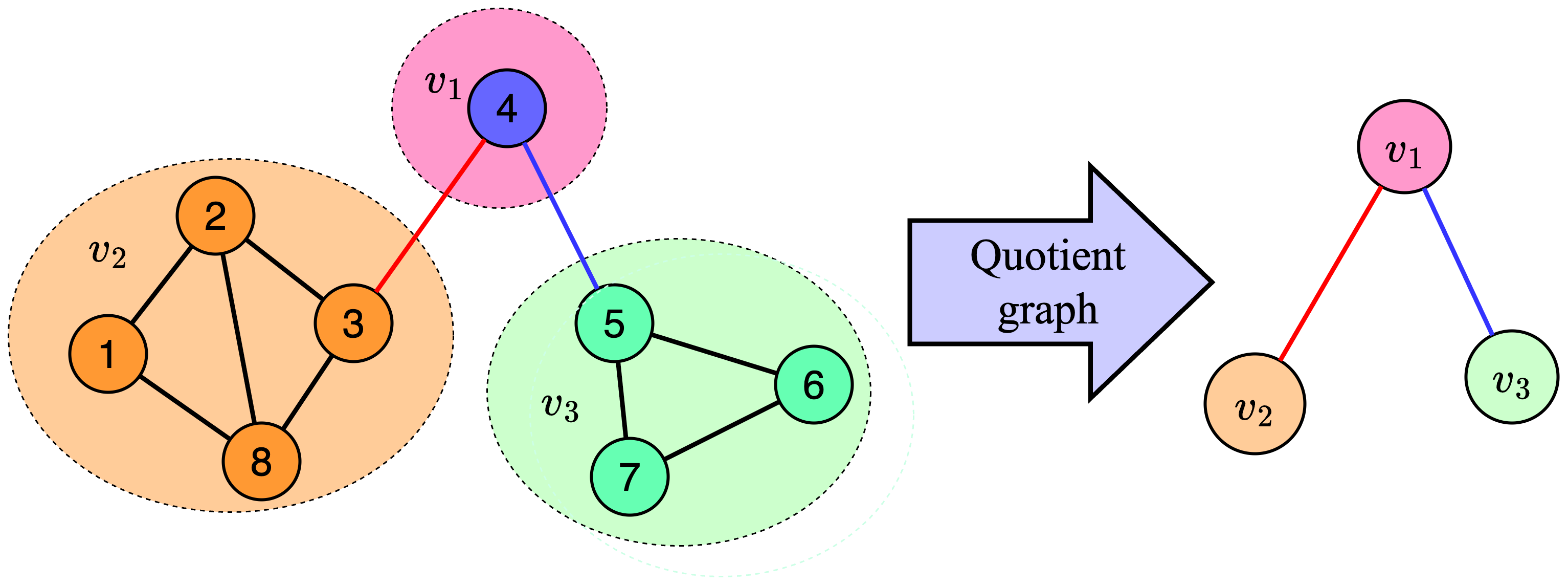}
    \caption{The right hand graph is the quotient graph under the equivalence relation specified by the partition set $\mathcal{G}_f / P = \{v_{1}, v_{2}, v_{3}\}$, where $V(G_{f}) = \{1, 2, ..., 8\}$, $v_{1} = \{4\}$, $v_{2} = \{1, 2, 3, 8\}$, and $v_{3} = \{5, 6, 7\}$.}
    \label{fig:quotient}
\end{figure}
%\begin{definition}

Let \( \mathcal{G}_f = (V_f, E_f) \) be a graph with node set \( V_f \) and edge set \( E_f \), and let \( P = \{V_1, V_2, \dots, V_k\} \) be a partition set of \( V_f \). The quotient graph~\cite{hajiabolhassan2023funqg} \( \mathcal{G}_q = \mathcal{G}_f / P = (V_q, E_q) \) has node set \( V_q = \{V_1, V_2, \dots, V_k\} \), where each node \( V_i \) represents a cluster of nodes from \( V_f \). Each node \( v \in V_f \) belongs to exactly one subset \( V_i \) in the partition, where \( i \in \{1,2,\dots,k\} \). This one-to-one assignment naturally induces an \textit{equivalence relation} \( \sim \) on \( V_f \), defined as following.
\begin{equation}
    v \sim w \quad \Longleftrightarrow \quad \exists \, V_i \in P \;\; \text{such that} \;\; v \in V_i \;\; \text{and} \;\; w \in V_i
\end{equation}

Fig.~\ref{fig:quotient} presents an example of how a quotient graph can be formed.

Let the landmark-level graph be defined as \( G = (V, E) \), where \( V \) is the set of landmark nodes and \( E \) is the set of landmark edges. The landmark nodes are partitioned into \( k \) disjoint regions using $k$-Means clustering based on their coordinates.

\begin{equation}
  \mathcal{P} = \{ R_1, R_2, \dots, R_k \}, \quad R_k \subseteq V, \quad R_i \cap R_j = \varnothing \ \text{for } i \neq j  
\end{equation}

The quotient graph is defined as \( G_q = (V_q, E_q) \), with \( |V_q| = k \), where each node \( u_k \in V_q \) corresponds to a region \( R_k \). The embedding of a region node is computed by mean pooling over the embeddings of its constituent landmarks:
\begin{equation}
\mathbf{h}_{R_k} = \frac{1}{|R_k|} \sum_{v_i \in R_k} \mathbf{h}_i
\end{equation}
Similarly, the coordinates of the region node are aggregated:
\begin{equation}
\mathbf{p}_{R_k} = \frac{1}{|R_k|} \sum_{v_i \in R_k} \mathbf{p}_i
\end{equation}

Edges in the quotient graph are defined using $k$-nearest neighbors ($k$-NN), based on the Euclidean distances between region coordinates \( \mathbf{p}_{R_k} \), ensuring connectivity reflects the spatial proximity of facial regions. In practice, this process is applied independently to each graph, producing a quotient graph with $k$ nodes per input graph.

The quotient graph is processed with two EdgeConv layers, as described in \eqref{eq:edgeconv}. These layers enrich the region embeddings by capturing higher-order dependencies, such as correlations between the eyes and mouth or between the eyebrows and cheeks.  
The refined embeddings improve the effectiveness of the downstream classification.
\subsection{Graph-Level Prediction from Region Embeddings}
To produce a unified representation of the facial expression for classification, the region embeddings of the quotient graph are processed to capture inter-region interactions and aggregated into a permutation-invariant graph-level embedding. This phase ensures that the model captures unique facial patterns, such as coordinated movements of the eyes and mouth, critical for distinguishing expressions like happiness or sadness in FER.

The refined region embeddings are aggregated via global additive pooling~\cite{kipf2017semisupervisedclassificationgraphconvolutional} to obtain a permutation-invariant graph-level representation:
\begin{equation}
    \mathbf{h}_G = \sum_{R \in V_q} \mathbf{h}_{R}
\end{equation}
Additive pooling ensures robust aggregation across regions, accommodating variations in region contributions across samples.

A linear layer maps \( \mathbf{h}_G \) to a probability distribution over 8 expression classes (neutral, happy, sad, surprise, fear, disgust, anger, contempt), enabling the model to classify facial expressions effectively.

\subsection{Theoretical Achievements}

The hierarchical architecture of the proposed GLaRE model achieves both computational efficiency and improved generalization for facial expression recognition by incorporating a quotient graph to represent regional facial structures. This approach reduces the computational complexity of message passing while maintaining the essential expressive information contained in facial landmarks.  

By coarsening the fine-level facial landmark graph into a higher-level quotient graph, the number of nodes and edges involved in computation is significantly reduced. Given a fine-level graph with \( N \) nodes and \( E \) edges, GNN-based message passing typically incurs \( \mathcal{O}(E) \) time complexity per layer. After coarsening, the coarse graph with \( N' \ll N \) nodes and \( E' \ll E \) edges reduces the message passing complexity to \( \mathcal{O}(E') \), which leads to significant speedup.

Beyond efficiency, the quotient graph groups landmarks into semantically meaningful facial regions (such as eyes, mouth, or eyebrows), thereby capturing higher-order interactions that are crucial for distinguishing expressions. This regional abstraction improves robustness to variations in landmark positions and local noise, enhancing the generalization ability of the model across diverse facial configurations.  

Finally, a permutation-invariant pooling mechanism aggregates region-level embeddings into a compact graph-level representation, ensuring consistent predictions irrespective of the input node ordering. This hierarchical approach allows the model to jointly exploit fine-grained landmark relations and coarse regional structures, resulting in a more efficient and reliable solution for facial expression recognition.

\section{Result \& Discussion}
This section presents the experimental evaluation of the proposed model GLaRE, which employs a hierarchical quotient graph to achieve computational efficiency and robustness to variations in facial landmark configurations. We provide a detailed analysis and discussion of classification accuracy, loss function, and comparisons with state-of-the-art models, including CNN-based approaches and graph-based methods, in this section. 

\subsection{Dataset Description}
We have conducted experiments on two benchmark facial expression datasets, AffectNet and FERG-DB, covering both real-world and stylized facial data.
\par \textbf{AffectNet}~\cite{mollahosseini2017affectnet} has emerged as the largest facial expression dataset to date, containing over one million facial images collected from the internet using emotion-related keywords in six different languages. Out of these, approximately 450,000 images have been manually annotated. The dataset defines eleven categories, including six basic emotions (anger, disgust, fear, happiness, sadness, surprise), as well as neutral, contempt, none, uncertain, and non-face. For our experiments, we have randomly created a subset consisting of the six basic expressions along with neutral, yielding a total of 83,901 training images and 1500 validation images.
\par \textbf{FERG-DB}~\cite{aneja2016modeling} (Facial Expression Research Group Database) consists of 555,767 images of six stylized characters exhibiting seven types of facial expressions: the six basic emotions and neutral. Since the facial figures in this dataset are cartoon-style and lack realistic facial texture and contour details, extracting accurate facial landmarks has been particularly challenging. This difficulty arises because many landmark detection models are trained on real human faces and often fail to generalize to synthetic domains.
\subsection{Baseline Methods}
The gACNN~\cite{li2018occlusion} proposes an occlusion-aware attention mechanism that combines local and global facial features to improve facial expression recognition under occluded conditions. LDL-ALSG~\cite{chen2020label} introduces soft label supervision and models the correlations between samples to enhance the learning of facial expression features. OADN~\cite{ding2020occlusion} incorporates a landmark-guided attention branch that directs the model to focus on non-occluded facial regions, thereby improving recognition performance under occlusions. FERGCN~\cite{liao2022fergcn} employs a deep graph-based architecture that includes a feature extraction module, a graph convolution module, and a graph matching mechanism to recognize facial expressions.
\subsection{Additional Details}
The model has been trained using the PyTorch-Geometric~\cite{fey2019fast} framework for 200 epochs. The Adam optimizer~\cite{kingma2014adam} has been employed with a learning rate of 0.0001 and default betas of $(0.9,\ 0.999)$. Cross-entropy loss has been used as the objective function, which is well-suited for multi-class classification tasks. It measures the discrepancy between the predicted class probabilities and the true labels, encouraging the network to assign higher confidence to the correct class. The loss for a single sample is defined as \eqref{eq:loss}.
\begin{equation}
    \lambda_{CE}=-\sum^{C}_{c=1}y_{c}log(\hat{y}_{c})
    \label{eq:loss}
\end{equation}
where $C$ is the total number of classes, $y_{c}$ is the ground truth indicator, and $\hat{y}_{c}$ is the predicted probability for class $c$.

\subsection{Performance Evaluation and Analysis}
\begin{table}[ht]
\renewcommand{\arraystretch}{1.3}
\setlength{\tabcolsep}{10pt} % adjusted for 4 columns
\centering
\caption{Emotion-wise accuracy and precision (with standard deviation) of the GLaRE model on AffectNet Dataset subset}
\begin{tabular}{lccc}
\hline
\textbf{Emotion} & \textbf{Samples per Class} & \textbf{Accuracy (\%)} & \textbf{Precision (\%)} \\
\hline
Fear     & 1190 & $65.29 \pm 0.82$ & $66.07 \pm 0.90$ \\
Anger    & 1236 & $65.13 \pm 0.76$ & $63.79 \pm 0.85$ \\
Joy      & 1157 & $65.25 \pm 0.79$ & $64.04 \pm 0.88$ \\
Sadness  & 1255 & $65.50 \pm 0.84$ & $66.18 \pm 0.91$ \\
Surprise & 1228 & $66.12 \pm 0.87$ & $66.56 \pm 0.92$ \\
Neutral  & 1218 & $64.94 \pm 0.80$ & $66.19 \pm 0.89$ \\
Disgust  & 1139 & $63.62 \pm 0.85$ & $63.07 \pm 0.93$ \\
\hline
\end{tabular}
\label{tab:emotion_accuracy}
\end{table}
\begin{figure}[ht!]
    \centering
    \includegraphics[width=\textwidth]{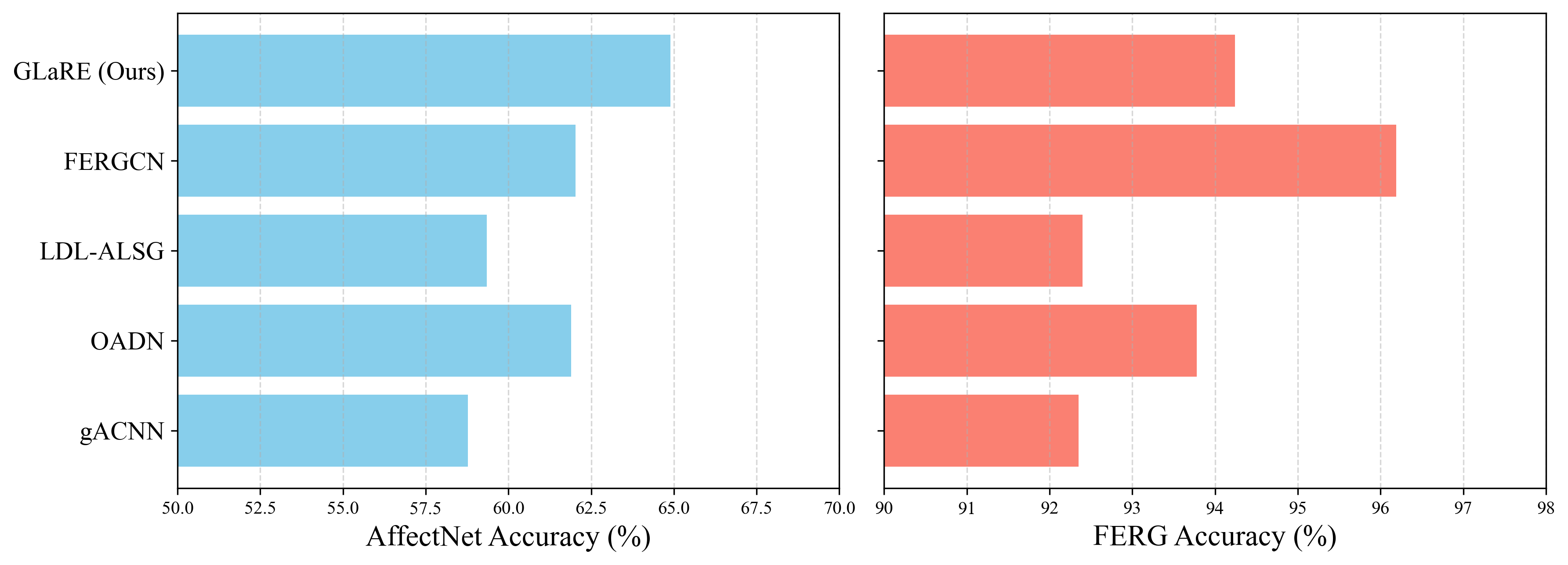}
    \caption{Model performance comparison on AffectNet and FERG datasets.}
    \label{fig:model_comparison}
\end{figure}
Table~\ref{tab:emotion_accuracy} summarizes the number of samples per class and their corresponding accuracies, while Fig.~\ref{fig:model_comparison} provides a visual representation of this performance comparison with baseline models. These results indicate that GLaRE is highly effective on real-world facial data, especially when landmark features are accurate and consistent.

\begin{table}[h!]
\renewcommand{\arraystretch}{1.3}
\setlength{\tabcolsep}{14pt} % adjust column separation
\centering
\caption{Accuracy (\%) comparison of various models on AffectNet and FERG datasets. GLaRE outperforms baselines on AffectNet and demonstrates competitive performance on FERG.}
\begin{tabular}{lcc}
\hline
\textbf{Method}   & \textbf{AffectNet} & \textbf{FERG} \\
\hline
gACNN             & 58.78$\pm$0.31     & 92.35$\pm$0.27 \\
OADN              & 61.89$\pm$0.28     & 93.78$\pm$0.34 \\
LDL-ALSG          & 59.35$\pm$0.25     & 92.40$\pm$0.29 \\
FERGCN            & 62.03$\pm$0.22     & \textbf{96.19$\pm$0.21} \\
\textbf{GLaRE}    & \textbf{64.89$\pm$0.23} & 94.24$\pm$0.38 \\
\hline
\end{tabular}
\label{tab:accuracy_comparison}
\end{table}
\begin{figure}[ht!]
    \centering
    \includegraphics[width=\textwidth]{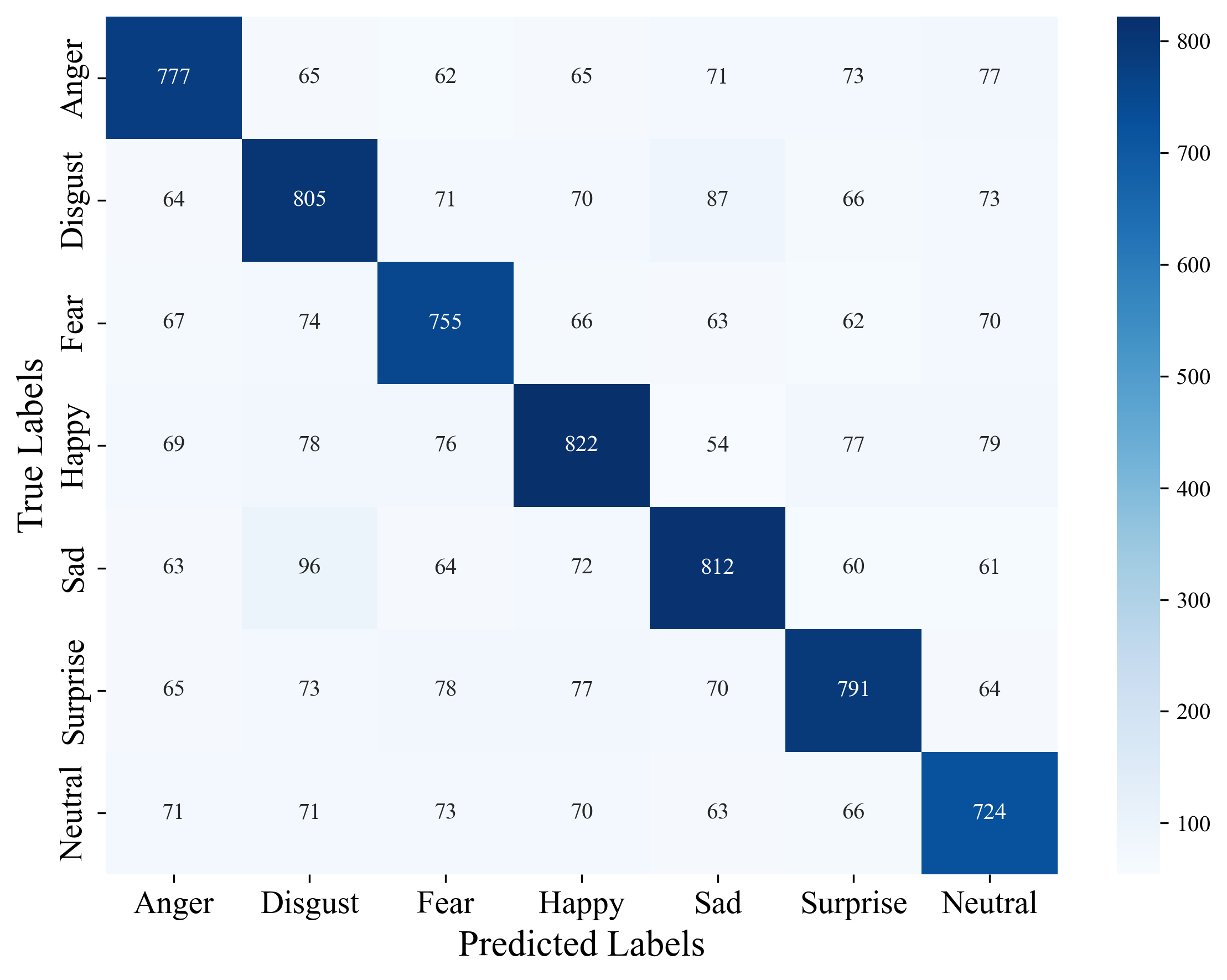}
    \caption{Confusion matrix of Emotion-wise accuracy on AffectNet Dataset.}
    \label{fig:confusion}
\end{figure}
As seen in Table \ref{tab:accuracy_comparison}, the proposed GLaRE model achieves the best accuracy of 64.89\% on this reduced AffectNet subset, outperforming all other baselines, including gACNN~\cite{li2018occlusion}, OADN~\cite{ding2020occlusion}, LDL-ALSG~\cite{chen2020label}, and FERGCN~\cite{liao2022fergcn}. To assess stability, ten independent runs have been conducted, and a low standard deviation of $\pm$0.27\% has been observed, confirming the consistency of the model’s performance. This demonstrates the model’s ability to generalize well, even with limited training data. Table \ref{tab:emotion_accuracy} further confirms that the performance remains consistent across different categories, with only minor variations. Performance comparisons have been conducted across all models on the same dataset, and statistical significance has been established through paired t-tests, yielding $p < 0.01$.

In contrast, while evaluating on the FERG dataset, which consists of synthetic or artificially-rendered characters, GLaRE achieves 94.24\% accuracy which is slightly lower than FERGCN’s 96.19\%. This reduction is primarily due to the difficulty in extracting consistent facial landmarks from stylized characters. Unlike real human faces, cartoon faces often lack anatomically grounded features, making landmark-based graph construction noisy and less reliable. Since GLaRE relies on high-quality 3D landmark features to build expressive graphs, its performance is naturally impacted under such conditions.

\par CNN-based models for FER often require millions of parameters, ranging from 2 to 18 million, which results in high computational demand. In contrast, graph-based approaches can achieve competitive results with far fewer parameters. The proposed GLaRE model contains only 44,424 parameters, highlights that GLaRE operates with significantly fewer parameters than established CNN-based methods while remaining more expressive than earlier graph-based baselines such as FERGCN~\cite{liao2022fergcn}. Such a lightweight design facilitates faster training, reduced memory consumption, and greater suitability for deployment in real-world scenarios.

\subsection{Ablation Studies}
\begin{figure}[ht!]
    \centering
    \includegraphics[width=0.9\linewidth]{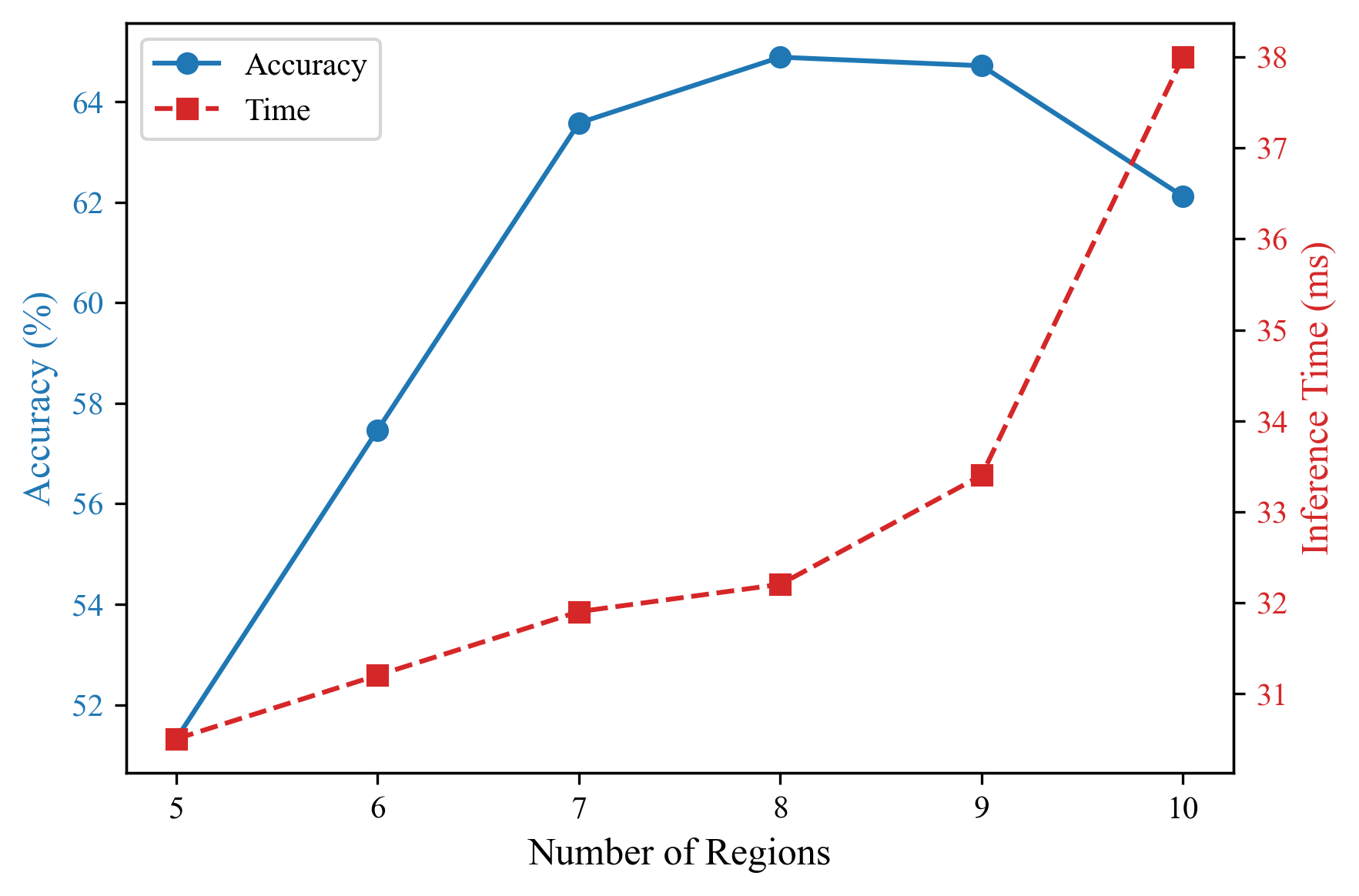}
    \caption{Ablation study on the number of quotient graph regions. The accuracy improves as the number of regions increases, peaking at 8–9 regions, beyond which performance slightly declines and the inference time increases.}
    \label{fig:ablation}
\end{figure}

We have conducted ablation studies to evaluate the contribution of the quotient graph and the choice of region granularity in our model. In the first setting, we have removed the quotient graph and directly processed landmark-level graphs. This has resulted in a significant increase in computational time and a noticeable drop in recognition accuracy, highlighting the efficiency and effectiveness of the quotient graph representation. Furthermore, we have varied the number of facial regions when constructing the quotient graph. The results presented in the Table~\ref{tab:ablation} and Fig.~\ref{fig:ablation} have shown that both decreasing and increasing the number of regions beyond a certain point leads to performance degradation. The highest accuracy has been achieved when the number of regions is set to eight or nine, suggesting that this range provides an optimal balance between structural compression and information preservation.

\begin{table}[h]
\centering
\caption{Ablation study on the number of regions in the quotient graph. Accuracy (\%) and inference time per batch (ms) are reported.}
\begin{tabular}{ccc}
\hline
\textbf{Number of Regions} & \textbf{Accuracy (\%)} & \textbf{Time (ms)} \\
\hline
5  & 51.32$\pm$0.28 & 30.5 \\
6  & 57.47$\pm$0.31 & 31.2 \\
7  & 63.58$\pm$0.26 & 31.9 \\
8  & \textbf{64.89$\pm$0.23} & 32.2 \\
9  & \textbf{64.72$\pm$0.25} & 33.4 \\
10 & 62.12$\pm$0.29 & 38.0 \\
\hline
\end{tabular}
\label{tab:ablation}
\end{table}
% \subsection{Ablation Studies on Featurization}

\par Additional ablation studies have been conducted to evaluate the contribution of different feature types. Two restricted variants of the input have been examined: (i) using only the 3D position vector of facial landmarks, and (ii) using only the appearance feature vector of size 16 extracted from local patches around the landmarks. As shown in Table~\ref{tab:ablation-features}, both variants have resulted in significantly lower performance compared to the joint featurization, demonstrating that the positional and appearance cues are complementary in nature. The position-only model has achieved $33\%$ accuracy, but it lacks the texture details necessary for reliable expression discrimination. Conversely, the appearance-only model has yielded only $56\%$ accuracy, highlighting that geometric information plays a crucial role in stabilizing the representation. In contrast, the joint featurization has achieved the best performance, confirming the necessity of integrating both modalities.

\begin{table}[ht!]
\centering
\caption{Ablation study on different feature choices for facial expression recognition. The results indicate that joint featurization of position and appearance vectors has provided a substantial improvement over using either feature alone.}
\label{tab:ablation-features}
\begin{tabular}{l c}
\toprule
\textbf{Feature Type} & \textbf{Accuracy (\%)} \\
\midrule
Position vector only (3D) & 56 \\
Appearance vector only (16D) & 33 \\
Joint (Position + Appearance) & \textbf{64.89} \\
\bottomrule
\end{tabular}
\end{table}

\section{Conclusion \& Future Work}
To conclude, GLaRE has been introduced as a hierarchical graph-based model that integrates fine-grained landmark information with high-level regional interactions for emotion recognition. The model has effectively captured both local geometry and global facial structure, while maintaining computational efficiency through graph coarsening. The approach has achieved performance surpassing existing baselines on a balanced subset of AffectNet and has demonstrated strong generalization on FERG-DB, despite the challenges posed by stylized facial textures.

Future work has been directed toward improving landmark detection on non-realistic faces, extending the framework to spatio-temporal modeling for video-based recognition, and adopting self-supervised strategies to reduce reliance on extensive annotated datasets.
% \section*{Code Availability Section}
% The code is available at \url{https://github.com/gnnplayground/GLaRE}
% \bibliographystyle{splncs04}
% \bibliography{main}

\end{document}